\documentclass{article}
\usepackage{ijcai22}

\usepackage{times}
\usepackage{soul}
\usepackage{url}
\usepackage[hidelinks]{hyperref}
\usepackage[utf8]{inputenc}
\usepackage[small]{caption}
\usepackage{graphicx}
\usepackage{amsmath}
\usepackage{amsthm}
\usepackage{booktabs}
\usepackage{algorithm}
\usepackage{algorithmic}
\usepackage{xcolor}
\title{Deep-Learning vs Regression: Prediction of Tourism Flow with Limited Data}

\author{
Julian Lemmel$^{\star 1, 2}$
\and
Zahra Babaiee$^{\star 1, 2}$
\and
Marvin Kleinlehner$^1$
\and
Ivan Majic$^3$\and
Philipp Neubauer$^{1}$\and
Johannes Scholz$^3$\and
Radu Grosu$^2$
\And
Sophie A. Neubauer (née Gruenbacher)$^{1,2}$
\affiliations
$^1$DatenVorsprung GmbH\\
$^2$TU Wien\\
$^3$TU Graz\\
$\star$ contributed equally
\emails
team@datenvorsprung.at
}

\begin{document}

\maketitle

\begin{abstract}
    
\end{abstract}
Modern tourism in the 21st century is facing numerous challenges. One of these challenges is the rapidly growing number of tourists in space limited regions such as historical city centers, museums or geographical bottlenecks like narrow valleys.
In this context, a proper and accurate prediction of tourism volume and tourism flow within a certain area is important and critical for visitor management tasks such as visitor flow control and prevention of overcrowding. Static flow control methods like limiting access to hotspots or using conventional low level controllers could not solve the problem yet.
In this paper, we empirically evaluate the performance of several state-of-the-art deep-learning methods in the field of visitor flow prediction with limited data by using available granular data supplied by a tourism region and comparing the results to ARIMA, a classical statistical method. 
Our results show that deep-learning models yield better predictions compared to the ARIMA method, while both featuring faster inference times and being able to incorporate additional input features.

\section{Introduction}

With increasing population and travel capacities (e.g. easy access to international flights) cultural tourism destinations have seen a rise in visitor counts. In addition, recent needs for social distancing and attendance limitations due to the global COVID-19 pandemic have confronted tourism destinations with significant challenges in preventing e.g. overcrowded waiting-lines. The perceptions of tourists regarding health hazards, safety and unpleasant tourism experiences may be influenced by social distance and better physical separation~\cite{SIGALA2020312}. Therefore, future-oriented tourism regions aim to control visitor flows in order to maximise visitor satisfaction, which is directly connected to the economical wealth of the specific tourism region.

 Unfortunately, many real-world problems suffer from limited data availability due to data compliance issues, lack of data collection or even lack of data transfer through stakeholders. In the end there are not enough datasets to properly train state-of-the art machine learning models. Since this is a generic problem, this scientific work is focusing on what is possible to achieve in the given environment considering the given data and data history.

This paper illuminates the work in progress of austrian based startup- and university-research. The first step in order to control tourist flows is to predict authentic movement and behavior patterns. However, since the tourist visitor flow is affected by many factors such as the weather, cultural events, holidays, and regional traffic, it is a very challenging task to accurately predict the future flow~\cite{DBLP:journals/corr/abs-1809-00101}. Due to the availability of large datasets and computational resources, deep neural networks became the state-of-the-art methods in the task of forecasting time-series data~\cite{ijcai2021-397}, including tourism flow applications~\cite{9271732}.

\begin{figure*}
    \centering
    \includegraphics[width=0.9\textwidth]{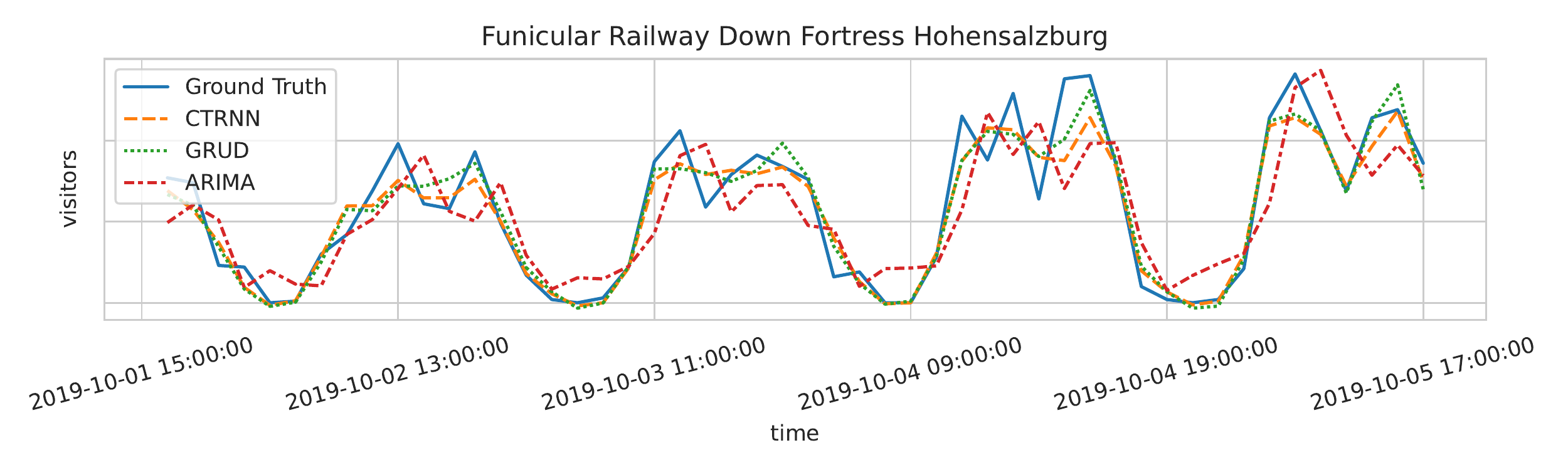}
    \includegraphics[width=0.9\textwidth]{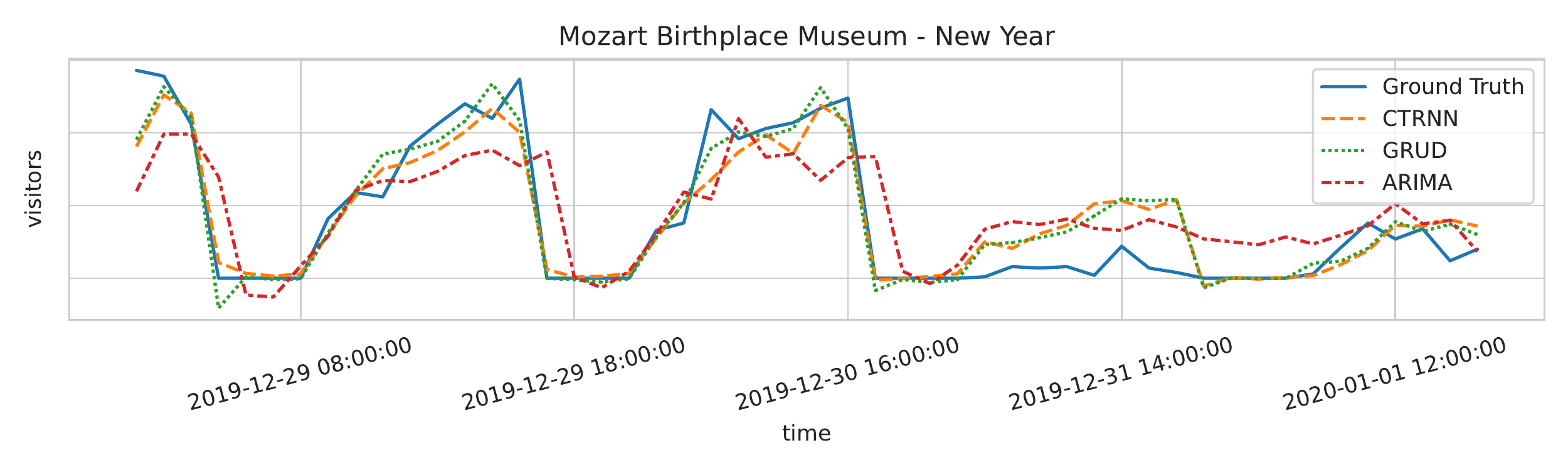}
    \caption{Predicted and True visitor counts for the Funicular Railway (top) and Mozart's Birthplace Museum (bottom). Predictions are computed using CT-RNN (orange), GRU-D (green) and ARIMA (red).}
    \label{fig:Bahn_Geburthaus}
\end{figure*}

In this work, we focus on tourist flow prediction based on a local dataset from the visitors of the tourist attractions of the city of Salzburg. After data preprocessing and dataset preparation, we attempt to compare the performance of different deep-learning based methods for time-series prediction with ARIMA, a traditional statistics based method. According to Li and Cao~\cite{liPredictionTourismFlow2018}, ARIMA is the most popular classical time forecasting method based on exponential smoothing and it was made popular in 1970s when it was proposed by Ahmed and Cook~\cite{ahmedAnalysisfreewaytraffic1979} to be used for short-term freeway traffic predictions. 

Deep neural networks are proven to work very well on large datasets. However, their performance can degrade when trained on limited data, resulting in poor predictions on the test set. Since limited data is a common problem in tourism time-series forecasting, we perform a comprehensive comparison of the DNNs and traditional techniques on a small dataset to reveal the shortcomings and point out necessary future improvements.


\textbf{Deep-learning based models.} Recurrent Neural Networks (RNNs) are the state-of-the-art models for learning time-series datasets. RNNs equip the neural networks with memory, making them successful at predicting the sequence-based data. The introduction of gating mechanism to RNNs lead to the great performance of LSTM~\cite{lstm} and GRU~\cite{GRU} networks. 

RNNs have limitations when facing irregularly-sampled time-series, present at many real-world forecasting problems such as tourist flow prediction. To address this limitation, phased-LSTM~\cite{phased_LSTM} adds a time gate to the LSTM cells. GRU-D~\cite{GRUD} incorporates time intervals by a trainable decaying mechanism in order to handle time-series with missing data and long-term dependencies.

Another approach is to introduce time-continuous models with latent state defined at all times such as CT-RNN~\cite{ctrnn}, CT-LSTM~\cite{CTLSTM} and CT-GRU~\cite{CTGRU}. A family of continuous-time networks are NeuralODEs~\cite{NODE} that define the hidden state of the network as a solution to an ordinary differential equation. Some limitations of NeuralODEs such as non-intersecting trajectories can be aleviated by using augmentations strategies leading to Augmented-NeuralODEs (ANODEs)~\cite{ANODE}. Continuous-time models share some favourable properties: Adaptive computation as they can be implemented by numerical ODE (ordinary differential equations) solvers and training with constant memory cost by using the adjoint sensitivity method~\cite{NODE}. In addition,  they can be statistically verified by using GoTube~\cite{GoTube2022} which constructs stochastic reachtubes (=the set of all reachable system states) of continuous-time systems and is made deliberately for the verification of countinuous-depth neural networks. 

\textbf{Traditional techniques.} For time-series forecasting with traditional techniques we use the Autoregressive Integrated Moving Average (ARIMA) model.
ARIMA has been used in recent studies as a baseline for the evaluation of novel deep-learning based models~\cite{yaoNeuralNetworkEnhanced2020,biDailyTourismVolume2020,liPredictionTourismFlow2018,hassaniForecastingAccuracyEvaluation2017} and is thus selected as a baseline model for this paper as well.

Since limited data is a common problem in tourism time-series forecasting, we summarize the specific contributions of our paper as follows:
\begin{itemize}
    \item  We perform a comprehensive comparison of the DNNs and ARIMA, a traditional technique, on a real-world dataset to reveal the shortcomings and point out necessary future improvements. 
    \item The real-world dataset is considered small because of limited historical entries.
    \item Per point-of-interest (POI), we perform granular predictions on an hourly basis, which is critical for the task of tourism flow control, see Figure~\ref{fig:Bahn_Geburthaus}.
    \item To the best of our knowledge, we are the first to apply a wide range of RNN models to tourist flow prediction.
\end{itemize}
\section{Related Work}
\begin{table}[ht]
    \centering
    \begin{tabular}{ |l c| }
    \toprule
     \textbf{Hyperparameter} & \textbf{Value} \\
     \hline
     sequence length & 30\\
     batch size & 16\\
     epochs & 300\\
     \hline
     optimizer & adam \\
     Learning-rate & $1e^{-3}$ \\  
     $\beta_{1,2}$ & $(0.9, 0.999)$\\    
     $\epsilon$ & $1e^{-8}$ \\
     \hline
     loss function & mse, mae, huber\\
     model size & 32, 64, 128\\
     normalized visitors & True, False\\
     \bottomrule
    \end{tabular}
    \caption{Hyperparameters used in RNN training.}
    \label{table:Hyperparameters}
\end{table}
Due to the importance of tourist flow prediction in the ever growing tourism industry, visitor forecasting has gained some attention over the past years~\cite{burgerPractitionersGuideTimeseries2001,hassaniForecastingAccuracyEvaluation2017}. 
Existing work examines tourist demand forecasting using Recurrent Neural Networks such as LSTMs ~\cite{7905712,liPredictionTourismFlow2018} or hidden Markov Models in conjunction with deep neural networks ~\cite{yaoNeuralNetworkEnhanced2020}. Most of these studies make predictions with only a limited set of models.

Data granularity is another important aspect of tourism data. Many studies focus on long-term predictions of monthly, quarterly and yearly, or in the best case daily number of visitors~\cite{biDailyTourismVolume2020} in large regions as city or country-level tourism demand~\cite{kursor}. However, performing granular predictions on an hourly basis and per POI is critical for the task of tourism flow control. 


\section{Methods}
For this work, we built our own dataset on hourly data collected from tourist attractions. We then perform predictions with a rich set of models and do a comprehensive comparison of the results.
In this section we first introduce the dataset we used for the experiments. Then we go over the methods we chose to evaluate and compare their performances.

\subsection{Data}
The dataset we used stems from the "Salzburg Card" which was kindly provided to us by TSG Tourismus Salzburg GmbH. Upon purchase of these cards, the owner has the ability to enter 32 different tourist attractions and museums included in the portfolio of the Salzburg Card. The dataset consists of the time-stamps of entries to each POI. Additionally, we used data about weather and holidays in Austria. 



\subsection{Deep-Learning models}
We use a large set of RNN variations on the tourist-flow dataset to perform a comprehensive comparison of the state-of-the art models and provide insight on their performance. The set comprises vanilla-RNN, LSTM, phased-LSTM, GRU-D, CT-RNN, CT-LSTM and Neural-ODE networks.

\subsection{Traditional methods}
In this study we use the non-seasonal variant of the ARIMA model which does not consider the seasonal patterns in a time-series. This model is usually denoted as \emph{ARIMA ($p$,$d$,$q$)}, where $p$ is the number of autoregressive terms, $d$ is the number of non-seasonal differences, and $q$ is the number of lagged forecast errors in the prediction equation~\cite{burgerPractitionersGuideTimeseries2001}.
A generic expression for the non-seasonal ARIMA process is given as~\cite{hyndmanAutomaticTimeSeries2008}:
\[ \phi(B)(1 - B^d)y_t = c + \theta(B)\varepsilon_t \]
\noindent where \{$\varepsilon_t$\} is a white noise process, $B$ is the backshift operator, and $\phi(z)$ and $\theta(z)$ are polynomials of order p and q respectively. A more detailed explanation of the ARIMA method and its working principles can be found in~\cite{hyndmanForecastingPrinciplesPractice2018}.

\section{Main Results}
In this section we describe our pipeline comprising pre-processing of the data, building different models and choosing their hyper-parameters.

\subsection{Data Preprocessing}
We used the Salzburg card data from years 2017, 2018, and 2019 for our experiments. In order to create the time-series data, we accumulated the hourly entries to each location. The data then consists of the hour of the day, and the number of entries at that hour to each of the 32 POIs. 

For the RNN models, we added additional features to the dataset: Year, Month, Day of month, Day of week, Holidays and Weather data. For the Holiday data we used the national holidays and school holidays and count the days to the next school day. For the Weather data, we use the hourly weather data with these features: Temperature, Feels Like, Wind speed, Precipitation, and Clouds as well as a One-Hot-Encoded single word description of the weather (e.g. "Snow").

We performed further pre-processing by normalizing all features to values between $0$ and $1$. To account for seasons, we performed sine-cosine transformation for the month. Intuitively, since it is a circular feature we do not want to have the values for December and January to be far from each other. 

Finally, We split the data into sequences of length 30, and used the data from years 2017 and 2018 as the training set, and 2019 as the test set.

\begin{table*}[ht]
  \centering
  \begin{tabular}{lcccc|cc|cc}
    \toprule
        &&&Training time& Prediction time& \multicolumn{2}{|c|}{only visitors}&\multicolumn{2}{c}{external features}\\
    \textbf{Model}     & \# Cells    & \# Parameters & (minutes) & (milliseconds) & MAE & RMSE & MAE & RMSE \\

    \midrule
    ARIMA & - & 224 & - & 69k & 5.217 & 7.833 & - & -\\
    \midrule
    ANODE           &    64.0 &       21.3k     &       145.6   & 3.01 & 4.599 & 6.965 &         4.410   &       6.663   \\
    Vanilla RNN     &   128.0 &       43.7k     &       5.9     & 0.18 & 3.958 & 6.321 &         3.802   &       6.160   \\
    LSTM            &    32.0 &       11.9k     & \textbf{1.5}  & 0.24 & 3.713 & 6.209 &         3.630   &       6.113   \\
    Phased LSTM     &    32.0 & \textbf{11.8k}  &       27.0    & 0.46 & 3.825 & 6.359 &         3.651   &       6.120   \\
    CT-LSTM         &    32.0 &       19.9k     &       18.1    & 0.31 & 3.734 & 6.239 &         3.700   &       6.185   \\
    CT-RNN          &   128.0 &       27.4k     &       57.1    & 0.60 & 3.694 & 6.131 &         3.629   & \textbf{5.983} \\
    GRU-D           &    64.0 &       27.7k     &       16.6    & 0.33 & 3.638 & 6.121 &  \textbf{3.621} &       6.073   \\

    \bottomrule
  \end{tabular}
  \caption{The prediction results for models when using only the visitors count data and when using additional features from external data. Our experimental results show that the errors were smaller for RNNs compared to ARIMA both with and without additional features were used.}
  \label{table:results}
\end{table*}

\subsection{ARIMA}
ARIMA parameters $p$, $d$, and $q$ define how the function will fit the given time-series which directly affects the quality of future predictions. There are several recommended approaches for the manual selection of parameters and they all rely on comparing the ARIMA model fit to the real values of the time-series to try and minimize the deviation between them~\cite{ahmedAnalysisfreewaytraffic1979,hyndmanAutomaticTimeSeries2008,hyndmanForecastingPrinciplesPractice2018}.

For this study, we have used the \emph{auto.arima} function from the \emph{R forecast} library that automatically fits the ARIMA model with different sets of parameters and returns the best combination of $p$, $d$, and $q$ for the given time-series~\cite{hyndmanAutomaticTimeSeries2008,hyndmanForecastingPrinciplesPractice2018}. Given that each of the 32 POIs in this study has a different time-series of visitor counts, we have determined different ARIMA parameters for each POI which were the best fit for its time-series.
ARIMA predictions models were then built individually for each POI by fitting the ARIMA to that POI's training dataset with the best $p$, $d$, and $q$ parameters for that POI using the Python \emph{pmdarima} library.
Each time after the number of visitors is predicted for the next hour in the test data, the true value (i.e., number of visitors) for that hour is added to update the existing ARIMA model and make it aware of all previous true values before making the next prediction.

\begin{figure*}
    \centering
    \includegraphics[width=0.9\textwidth]{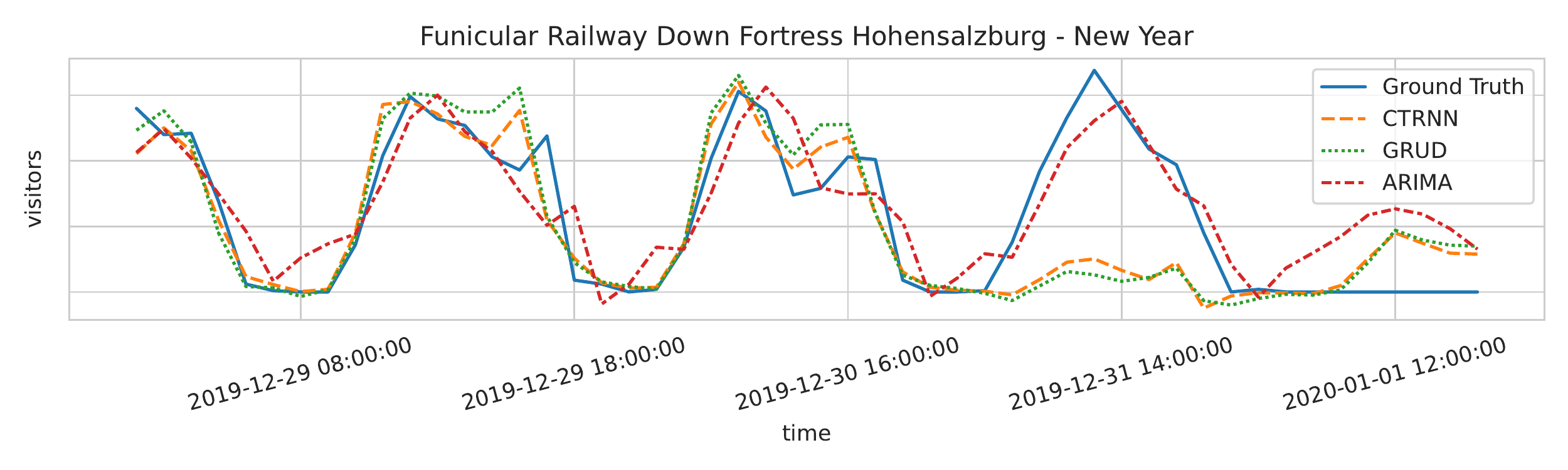}
    \includegraphics[width=0.9\textwidth]{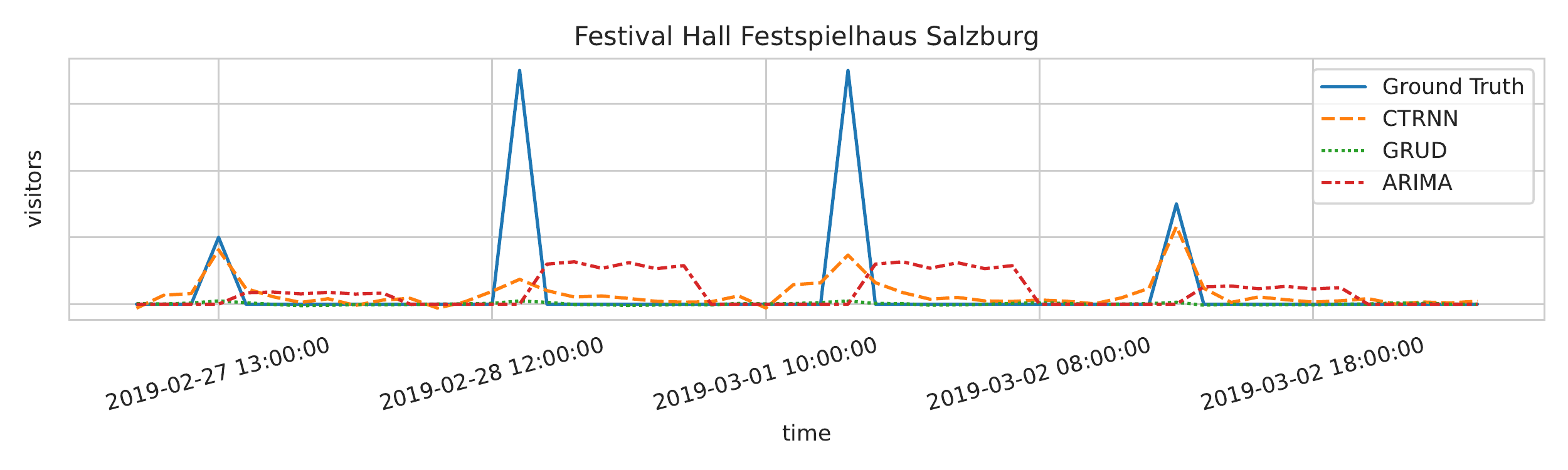}
    \caption{Predicted and True visitor counts for the Funicular Railway around New Year's Eve (top) and the Festival Hall (bottom).}
    \label{fig:BahnNYE_Festspielhaus}
\end{figure*}

\subsection{RNN training}

All of the Neural Networks were trained with Backpropagation-Through-Time and the Adam optimizer \cite{kingma2014method} using the parameters given in Table~\ref{table:Hyperparameters}. In order to find optimal model size, loss function, and whether to use normalized visitor counts, we did a grid search conducting three runs per configuration and keeping the one which achieved the lowest RMSE. The loss functions listed in Table~\ref{table:Hyperparameters} correspond to Mean-Squared-Error, Mean-Absolute-Error and Huber-Loss respectively.

\subsection{Comparing ARIMA with Neural Networks}

Before proceeding to the empirical comparison of the predictions made by each model, we mention some of the advantages of the deep-learning (DL) models over the ARIMA model. 
\begin{itemize}
    \item ARIMA can not handle additional features and makes predictions merely on the number of visitors in the past. The DL models were equipped with additional features of time, weather and holidays.
    \item ARIMA fails when given only short sequences as input. The DL models can handle shorter sequences when trained on the whole training set. To ensure fairness, we feed the whole past sequence to ARIMA when evaluating at each time-step.
    \item ARIMA makes predictions for each POI separately, while the DL models are trained with the visitors to all POIs in a single vector and make predictions for all at the same time. Thus, ARIMA loses the implicit data about the total number of visitors in the city.
    \item ARIMA takes a lot of time for the prediction of each time-step and each POI, while using the DL models requires only a one-time cost of training and afterwards the evaluation is fast.
\end{itemize}

\section{Experimental Evaluation}
We perform a diverse set of experiments with ARIMA and DL models to evaluate their forecasting accuracy, execution time and prediction time and compare the models. We run our evaluations on a standard workstation machine setup (4 CPU cores with 3.9 GHz, 8GB DDR3 memory) equipped with a single GPU (GeForce GTX 1050 Ti, 4GB memory).

Table~\ref{table:results} shows the Mean-Absolute-Error (MAE) and Root-Mean-Squared-Error (RMSE) achieved for each method. ARIMA was not able to use the external features. Hence, in order to ensure the fairness in our comparisons, we report the results with and without using the additional features of weather and holidays for DL models.     
For the deep-learning models model size, corresponding number of parameters and training and prediction times of the best run are listed additionally. Since using normalized visitor counts led to better results in every single model, we omitted non-normalized models from the table. All of the models except the ANODE achieved best results when trained with MAE as the loss function. For ANODE huber loss was best.
The phased LSTM uses the least number of parameters for comparable results.

All of our deep-learning models were able to outperform the ARIMA method in both metrics, with and without additional features. Providing the additional features to the models resulted in a slightly better performance for the DL models. However, the improvements were quite marginal over the results when only using the number of visitors. This can be because we have limited training data available and thus increasing the number of features does not pay off, or even might result in over-fitting.  

Interestingly, using MAE as loss function during training led to best results in the majority of cases. Since MAE is computed by summing up the absolute errors, it doesn't weigh outliers as strongly as MSE does. For the purpose of comparing the precision of the prediction, we used RMSE as it is very important to avoid huge differences between the actual visitors and the prediction for it. Small differences are not that relevant in the real world.

We measured training times for the DL models and prediction times for all models. As shown in Table~\ref{table:results} ARIMA took 69s to perform a single prediction for all POIs and the DL took for the same task fractions of milliseconds, while having used once y minutes for training. It is not possible to directly compare training and testing runtimes between them because ARIMA does not have a dedicated training step where the model is built. Instead, before every new prediction at time t, ARIMA model is fitted to the known time series up until the time t-1 (i.e., training and testing are integrated for every prediction). Therefore, our approach favors ARIMA accuracy over prediction time.
For this particular use case where predictions of the number of visitors are performed hourly with the latest information, we considered the runtime of a single prediction as the DL models only need to be trained once.

In order to visually explore the predictions made by the models, we plotted the predictions and the ground truth for a few selected time-windows. We plot the predictions made by the DL models (including the external features) with the best MAE and RMSE, which were the GRU-D and CT-RNN respectively. The prediction made by the DL models with the visitors only data was only slightly worse than the others, which is why we omit these evaluations in the plots. Our plots show that although ARIMA is out-performed by the DL methods in the average error of all predictions, there are cases where it actually performs better than the other models.

Figure~\ref{fig:Bahn_Geburthaus} top shows the forecast and real values for the tourists entered the Funicular Railway descend which is the cable car ride leading up to Salzburg Castle. As visible in the plot, the DL models show a better performance, especially in the valleys where the ARIMA fails to predict the downfalls accurately.

Figure \ref{fig:Bahn_Geburthaus} bottom shows visitor predictions for Mozart's Birthplace Museum around the time of New Year's Eve. The reduced numbers of visitors on the 1st and 2nd of January is expected but nonetheless overestimated by all our models.

Figure \ref{fig:BahnNYE_Festspielhaus} top again shows the data for the cable car but around New Year's Eve. Neither CT-RNN nor GRU-D were able to correctly predict the visitors in those days while ARIMA gets it right on the 31st of December. Apparently, the ride was closed on Jan 1st since the true visitor count stays at 0. None of our predictions saw that coming.

Figure~\ref{fig:BahnNYE_Festspielhaus} bottom shows the predcitions for the Festival Hall Festspielhaus tour which is with sparse entries since it takes place daily at 2 pm. All models fail in prediction for the second and third peak at this location. However, CT-RNN shows a very good performance in predicting the first and last peak and at least shows an upward trend for the second and third peak. ARIMA can not handle this type of sparse data at all.

\section{Conclusions and future work}
We performed a thorough evaluation of deep-learning (DL) versus ARIMA, a traditional method, on the task of forecasting tourist flow time-series. We found that all of the DL models were able to outperform the established ARIMA method when using only visitor counts as the training data. Extending the dataset with additional features of time, holidays, and weather improved the predictions of the DL models, while ARIMA is not able to handle additional features. While the improvements were little, the performance might still get boosted by performing feature selection for the weather data and improving the holidays features. In terms of the prediction time, DL models are meaningfully faster than ARIMA, which is an important aspect since most real-world applications require fast inference time.

This paper opens many avenues for future research. The most straightforward next step is to try to improve the performance of the DL methods using additional training techniques such as regularization or learning rate scheduling.
Admittedly, the ARIMA method is most suitable for predicting univariate data which is why our classical methods should be extended by also considering methods for multivariate forecasting such as Vector autoregression (VAR). Furthermore, there also exist online methods (e.g. Online ARIMA \cite{liuOnlineARIMAAlgorithms}) that allow for adjusting the model for new datapoints.

In this work we did not include the spatial data of the locations. Working on finding the best way to incorporate the geographical features and distances of the POIs is another future direction, since it's critical to include them while working on the tourism flow datasets. Another challenging direction is to attempt predictions on unfamiliar situations with limited data. We did not include the post-COVID years in this study, since there is a lot of anomalies in the tourism section in these years. We are confident however, that by incorporating inductive biases such as information about lockdowns imposed by the government our deep-learning models could produce useful predictions.
In addition, we want to use the knowledge gained to build specialised models which outperforms state-of-the art models in terms of short-term prediction with limited data. Another direction is to work on predictions for a longer time horizon and incorporate these predictions into a recommender system for tourists. Therefore we would go one step further into the direction of giving tourism regions the ability to control the visitor flow.





\bibliographystyle{named}
\bibliography{ijcai22}

\begin{thebibliography}{}

\bibitem[\protect\citeauthoryear{Ahmed and
  Cook}{1979}]{ahmedAnalysisfreewaytraffic1979}
Mohammed~S. Ahmed and Allen~R. Cook.
\newblock Analysis of freeway traffic time-series data by using box-jenkins
  techniques.
\newblock {\em Transportation Research Record}, (722), 1979.
\newblock ISBN: 9780309029728.

\bibitem[\protect\citeauthoryear{Asvikarani \bgroup \em et al.\egroup
  }{2020}]{kursor}
Ayu Asvikarani, I~Widiartha, and Made Raharja.
\newblock Foreign tourist arrival forecasting to bali using cascade forward
  backpropagation.
\newblock {\em Jurnal Ilmiah Kursor}, 10(4), 2020.

\bibitem[\protect\citeauthoryear{Bi \bgroup \em et al.\egroup
  }{2020}]{biDailyTourismVolume2020}
Jian-Wu Bi, Yang Liu, and Hui Li.
\newblock Daily tourism volume forecasting for tourist attractions.
\newblock {\em Annals of Tourism Research}, 83(102923):1--22, 2020.

\bibitem[\protect\citeauthoryear{Burger \bgroup \em et al.\egroup
  }{2001}]{burgerPractitionersGuideTimeseries2001}
C.~J. S.~C Burger, M~Dohnal, M~Kathrada, and R~Law.
\newblock A practitioners guide to time-series methods for tourism demand
  forecasting \textemdash{} a case study of {{Durban}}, {{South Africa}}.
\newblock {\em Tourism Management}, 22(4):403--409, August 2001.

\bibitem[\protect\citeauthoryear{Che \bgroup \em et al.\egroup }{2018}]{GRUD}
Zhengping Che, Sanjay Purushotham, Kyunghyun Cho, David Sontag, and Yan Liu.
\newblock Recurrent neural networks for multivariate time series with missing
  values.
\newblock {\em Scientific Reports}, 8(1), 2018.

\bibitem[\protect\citeauthoryear{Chen \bgroup \em et al.\egroup }{2018}]{NODE}
Tian~Qi Chen, Yulia Rubanova, Jesse Bettencourt, and David Duvenaud.
\newblock Neural ordinary differential equations.
\newblock {\em CoRR}, abs/1806.07366, 2018.

\bibitem[\protect\citeauthoryear{Chung \bgroup \em et al.\egroup }{2014}]{GRU}
Junyoung Chung, {\c{C}}aglar G{\"{u}}l{\c{c}}ehre, KyungHyun Cho, and Yoshua
  Bengio.
\newblock Empirical evaluation of gated recurrent neural networks on sequence
  modeling.
\newblock {\em CoRR}, abs/1412.3555, 2014.

\bibitem[\protect\citeauthoryear{Dupont \bgroup \em et al.\egroup
  }{2019}]{ANODE}
Emilien Dupont, Arnaud Doucet, and Yee~Whye Teh.
\newblock Augmented {{Neural ODEs}}.
\newblock {\em arXiv:1904.01681 [cs, stat]}, October 2019.

\bibitem[\protect\citeauthoryear{Gruenbacher \bgroup \em et al.\egroup
  }{2022}]{GoTube2022}
Sophie Gruenbacher, Mathias Lechner, Ramin Hasani, Daniela Rus, Thomas~A.
  Henzinger, Scott~A. Smolka, and Radu Grosu.
\newblock Gotube: Scalable statistical verification of continuous-depth models.
\newblock {\em AAAI}, May 2022.

\bibitem[\protect\citeauthoryear{Hassani \bgroup \em et al.\egroup
  }{2017}]{hassaniForecastingAccuracyEvaluation2017}
Hossein Hassani, Emmanuel~Sirimal Silva, Nikolaos Antonakakis, George Filis,
  and Rangan Gupta.
\newblock Forecasting accuracy evaluation of tourist arrivals.
\newblock {\em Annals of Tourism Research}, 63:112--127, March 2017.

\bibitem[\protect\citeauthoryear{Hochreiter and Schmidhuber}{1997}]{lstm}
Sepp Hochreiter and Jürgen Schmidhuber.
\newblock Long short-term memory.
\newblock {\em Neural Computation}, 9(8):1735--1780, 1997.

\bibitem[\protect\citeauthoryear{Hyndman and
  Athanasopoulos}{2018}]{hyndmanForecastingPrinciplesPractice2018}
Rob~J. Hyndman and George Athanasopoulos.
\newblock {\em Forecasting: {{Principles}} and {{Practice}} (2nd Ed)}.
\newblock {OTexts}, {Melbourne, Australia}, second edition, 2018.

\bibitem[\protect\citeauthoryear{Hyndman and
  Khandakar}{2008}]{hyndmanAutomaticTimeSeries2008}
Rob~J. Hyndman and Yeasmin Khandakar.
\newblock Automatic {{Time Series Forecasting}}: {{The}} forecast {{Package}}
  for {{R}}.
\newblock {\em Journal of Statistical Software}, 27(3):1--22, 2008.

\bibitem[\protect\citeauthoryear{ichi Funahashi and Nakamura}{1993}]{ctrnn}
Ken ichi Funahashi and Yuichi Nakamura.
\newblock Approximation of dynamical systems by continuous time recurrent
  neural networks.
\newblock {\em Neural Networks}, 6(6):801--806, 1993.

\bibitem[\protect\citeauthoryear{Kingma and Ba}{2014}]{kingma2014method}
Diederik~P. Kingma and Jimmy Ba.
\newblock Adam: A method for stochastic optimization, 2014.
\newblock cite arxiv:1412.6980Comment: Published as a conference paper at the
  3rd International Conference for Learning Representations, San Diego, 2015.

\bibitem[\protect\citeauthoryear{Li and
  Cao}{2018}]{liPredictionTourismFlow2018}
YiFei Li and Han Cao.
\newblock Prediction for {{Tourism Flow}} based on {{LSTM Neural Network}}.
\newblock {\em Procedia Computer Science}, 129:277--283, January 2018.

\bibitem[\protect\citeauthoryear{Liu \bgroup \em et al.\egroup
  }{}]{liuOnlineARIMAAlgorithms}
Chenghao Liu, Steven C~H Hoi, Peilin Zhao, and Jianling Sun.
\newblock Online {{ARIMA Algorithms}} for {{Time Series Prediction}}.
\newblock page~7.

\bibitem[\protect\citeauthoryear{Liu \bgroup \em et al.\egroup
  }{2018}]{DBLP:journals/corr/abs-1809-00101}
Lingbo Liu, Ruimao Zhang, Jiefeng Peng, Guanbin Li, Bowen Du, and Liang Lin.
\newblock Attentive crowd flow machines.
\newblock {\em CoRR}, abs/1809.00101, 2018.

\bibitem[\protect\citeauthoryear{Mei and Eisner}{2017}]{CTLSTM}
Hongyuan Mei and Jason~M Eisner.
\newblock The neural hawkes process: A neurally self-modulating multivariate
  point process.
\newblock In I.~Guyon, U.~Von Luxburg, S.~Bengio, H.~Wallach, R.~Fergus,
  S.~Vishwanathan, and R.~Garnett, editors, {\em Advances in Neural Information
  Processing Systems}, volume~30. Curran Associates, Inc., 2017.

\bibitem[\protect\citeauthoryear{Mozer \bgroup \em et al.\egroup
  }{2017}]{CTGRU}
Michael~C. Mozer, Denis Kazakov, and Robert~V. Lindsey.
\newblock Discrete event, continuous time rnns.
\newblock {\em CoRR}, abs/1710.04110, 2017.

\bibitem[\protect\citeauthoryear{Neil \bgroup \em et al.\egroup
  }{2016}]{phased_LSTM}
Daniel Neil, Michael Pfeiffer, and Shih-Chii Liu.
\newblock Phased lstm: Accelerating recurrent network training for long or
  event-based sequences.
\newblock In D.~Lee, M.~Sugiyama, U.~Luxburg, I.~Guyon, and R.~Garnett,
  editors, {\em Advances in Neural Information Processing Systems}, volume~29.
  Curran Associates, Inc., 2016.

\bibitem[\protect\citeauthoryear{Pan \bgroup \em et al.\egroup
  }{2021}]{ijcai2021-397}
Qingyi Pan, Wenbo Hu, and Ning Chen.
\newblock Two birds with one stone: Series saliency for accurate and
  interpretable multivariate time series forecasting.
\newblock In Zhi-Hua Zhou, editor, {\em Proceedings of the Thirtieth
  International Joint Conference on Artificial Intelligence, {IJCAI-21}}, pages
  2884--2891. International Joint Conferences on Artificial Intelligence
  Organization, 8 2021.
\newblock Main Track.

\bibitem[\protect\citeauthoryear{Prilistya \bgroup \em et al.\egroup
  }{2020}]{9271732}
Suci~Karunia Prilistya, Adhistya Erna~Permanasari, and Silmi Fauziati.
\newblock Tourism demand time series forecasting: A systematic literature
  review.
\newblock In {\em 2020 12th International Conference on Information Technology
  and Electrical Engineering (ICITEE)}, pages 156--161, 2020.

\bibitem[\protect\citeauthoryear{Rizal and Hartati}{2016}]{7905712}
Ahmad~Ashril Rizal and Sri Hartati.
\newblock Recurrent neural network with extended kalman filter for prediction
  of the number of tourist arrival in lombok.
\newblock In {\em 2016 International Conference on Informatics and Computing
  (ICIC)}, pages 180--185, 2016.

\bibitem[\protect\citeauthoryear{Sigala}{2020}]{SIGALA2020312}
Marianna Sigala.
\newblock Tourism and covid-19: Impacts and implications for advancing and
  resetting industry and research.
\newblock {\em Journal of Business Research}, 117:312--321, 2020.

\bibitem[\protect\citeauthoryear{Yao and
  Cao}{2020}]{yaoNeuralNetworkEnhanced2020}
Yuan Yao and Yi~Cao.
\newblock A {{Neural}} network enhanced hidden {{Markov}} model for tourism
  demand forecasting.
\newblock {\em Applied Soft Computing Journal}, 94(106465):1--20, September
  2020.

\end{thebibliography}

\end{document}